\documentclass{article}

\usepackage{microtype}
\usepackage{graphicx}
\usepackage{subfigure}
\usepackage{booktabs} 

\usepackage{hyperref}
\usepackage{gensymb}

\usepackage[accepted]{icml2024}

\usepackage{amsmath}
\usepackage{amssymb}
\usepackage{mathtools}
\usepackage{amsthm}
\usepackage{amsmath}

\DeclareMathOperator*{\argmin}{arg\,min}

\usepackage[capitalize,noabbrev]{cleveref}

\theoremstyle{plain}

\theoremstyle{definition}

\theoremstyle{remark}

\usepackage[textsize=tiny]{todonotes}

\icmltitlerunning{Chain-of-Thought Predictive Control}

\begin{document}

\twocolumn[
\icmltitle{Chain-of-Thought Predictive Control}




\begin{icmlauthorlist}
\icmlauthor{Zhiwei Jia}{ucsd}
\icmlauthor{Vineet Thumuluri}{ucsd}
\icmlauthor{Fangchen Liu}{ucb}
\icmlauthor{Linghao Chen}{zju}
\icmlauthor{Zhiao Huang}{ucsd}
\icmlauthor{Hao Su}{ucsd}
\end{icmlauthorlist}

\icmlaffiliation{ucsd}{UC San Diego}
\icmlaffiliation{ucb}{UC Berkeley}
\icmlaffiliation{zju}{Zhejiang University}

\icmlcorrespondingauthor{Zhiwei Jia}{zjia@ucsd.edu}

\icmlkeywords{Machine Learning, ICML}

\vskip 0.3in
]



\printAffiliationsAndNotice{}  

\begin{abstract}
We study generalizable policy learning from demonstrations for complex low-level control (e.g., contact-rich object manipulations). 
We propose a novel hierarchical imitation learning method that utilizes sub-optimal demos.
Firstly, we propose an observation space-agnostic approach that efficiently discovers the multi-step subskill decomposition of the demos in an unsupervised manner.
By grouping temporarily close and functionally similar actions into subskill-level demo segments, the observations at the segment boundaries constitute a chain of planning steps for the task, which we refer to as the \textit{chain-of-thought} (CoT).  
Next, we propose a Transformer-based design that effectively learns to predict the CoT as the subskill-level guidance. 
We couple action and subskill predictions via learnable prompt tokens and a hybrid masking strategy, which enable dynamically updated guidance at test time and improve feature representation of the trajectory for generalizable policy learning.
Our method, Chain-of-Thought Predictive Control (\textbf{CoTPC}), consistently surpasses existing strong baselines on challenging manipulation tasks with sub-optimal demos. See more details at our project \href{https://sites.google.com/view/cotpc}{page}.

\end{abstract}
 
\section{Introduction} \label{sec:intro}
Hierarchical RL (HRL) \cite{hutsebaut2022hierarchical} has attracted much attention in the AI community as a promising direction for sample-efficient and generalizable policy learning.
HRL tackles complex sequential decision-making problems by decomposing them into simpler and smaller sub-problems via temporal abstractions (the so-called Chain-of-Thought \cite{wei2022chain}).
In addition, many adopt a two-stage policy and possess the planning capabilities for high-level actions (i.e., subgoals or options) to achieve generalizability. 
On the other hand, imitation learning (IL) remains one of the most powerful approaches to training autonomous agents. 
Without densely labeled rewards or on-policy/online interactions, IL usually casts policy learning as (self-)supervised learning with the potential to leverage large-scale pre-collected demonstrations, usually with Transformer, as inspired by the recent success of large language models (LLMs).
An obstacle in building foundational decision-making models \cite{yang2023foundation} remains the better use of sub-optimal demonstrations.
In this paper, we study hierarchical IL from sub-optimal demonstrations for low-level control tasks.

Despite the recent progress \cite{chen2021decision,florence2022implicit,shafiullah2022behavior,liu2022masked,ajay2022conditional,chi2023diffusion}, it remains extremely challenging to solve low-level control tasks such as contact-rich object manipulations by IL in a scalable manner.
Usually, the demonstrations are inherently sub-optimal because of the underlying contact dynamics \cite{pfrommer2021contactnets} and the way they are produced.
The undesirable properties, such as being non-Markovian, noisy, discontinuous, and random, pose great challenges in both the optimization and the generalization of the imitators (see more in Appendix \ref{app:challenges1}).
We find that, by adopting the hierarchical principles (i.e., temporal abstraction and high-level planning) into our Transformer-based design, we can better leverage sub-optimal demos to solve challenging tasks.
To achieve this, we first propose an unsupervised subskill discovery strategy to generate subskill-level supervision from the demonstrations. 
We then train our model to dynamically generate subskill guidance for better low-level action predictions.

Specifically, we consider the \textit{multi-step subskill decomposition of a task into a chain of planning steps as its chain-of-thought}, CoT, inspired by \cite{wei2022chain,yang2022chain}.
As part one of our contribution, we propose an observation space-agnostic approach that efficiently discovers the subskills, defined as a sequence of key observations of the demos, in an unsupervised manner.
We propose to group temporarily close and functionally similar actions into a segment.
Then the change points (the segment boundaries) naturally constitute the subskill sequence, i.e., the CoT, that represents the high-level task completion process.
For part two, we propose a novel Transformer-based design that effectively learns to predict the CoT jointly with the low-level actions. 
This coupled prediction mechanism is achieved by adding additional prompt tokens at the beginning of the context history and by adopting a hybrid masking strategy.
As a result, the subskill guidance is dynamically updated at each step and better feature representation of the trajectories is learned, eventually improving the generalizability of the low-level action prediction process. 

Our method, Chain-of-Thought Predictive Control (CoTPC), learns faster from sub-optimal demos, from an optimization perspective, by utilizing the subskills that are usually more robust and admit less variance.
From a generalization perspective, it uses Transformers \cite{brown2020language} to improve generalization with subskill planning, which is learned with the unsupervisedly discovered CoTs as supervision.
We evaluate CoTPC on several challenging low-level control tasks (Moving Maze, Franka-Kitch and ManiSkill2) and verify its design with ablation studies.
We find that CoTPC consistently outperforms several strong baselines.

\section{Related Work}

We include additional related work in Appendix \ref{app:more_dis}.

\paragraph{Learning from Demonstrations (LfD)}
Learning interactive agents from pre-collected demos has been popular due to its effectiveness and scalability.
Roughly speaking, there are three categories: offline RL, online RL with auxiliary demos, and behavior cloning (BC).
While offline RL approaches \cite{kumar2019stabilizing,fu2020d4rl,levine2020offline,kumar2020conservative,kostrikov2021offline,chen2021decision,wang2022diffusion} usually require demonstration with densely labeled rewards and the methods that augment online RL with demos \cite{hester2018deep,kang2018policy,ross2011reduction,nair2020accelerating,rajeswaran2017learning,ho2016generative,pertsch2021accelerating,singh2020parrot} rely on on-policy interactions, BC \cite{pomerleau1988alvinn} formulates fully supervised or self-supervised learning problems with better practicality and is adopted widely, especially in robotics \cite{zeng2021transporter,florence2022implicit,qin2022one,zhang2018deep,brohan2022rt,rahmatizadeh2018vision,florence2019self,zeng2020tossingbot}.
However, a well-known shortcoming of BC is the compounding error \cite{ross2011reduction}, usually caused by the distribution shift between the demo and the test-time trajectories.
Various methods were proposed to tackle it \cite{ross2010efficient,ross2011reduction,sun2017deeply,laskey2017dart,tennenholtz2021covariate,brantley2019disagreement,chang2021mitigating}.
Other issues include non-Markovity \cite{mandlekar2021matters}, discontinuity \cite{florence2022implicit}, randomness and noisiness \cite{sasaki2020behavioral,wu2019imitation} of the demos 
that results in great compounding errors of neural policies during inference (see Appendix A for detailed discussions).

\paragraph{LfD as Sequence Modeling}
A recent trend in offline policy learning is to relax the Markovian assumption of policies, inspired by the success of sequence models \cite{graves2012long,chung2014empirical,vaswani2017attention} where model expressiveness and capacity are preferred over algorithmic sophistication.
Among these, \cite{dasari2021transformers,mandi2021towards} study one-shot imitation learning, \cite{lynch2020learning,singh2020parrot} explore behavior priors from demos, \cite{chen2021decision,liu2022masked,janner2021offline,shafiullah2022behavior,ajay2022conditional,janner2022planning} examine different modeling strategies for policy learning.
In particular, methods based on Transformers \cite{vaswani2017attention,brown2020language} are extremely popular due to their simplicity and effectiveness.

\paragraph{Hierarchical Approaches in Sequence Modeling and RL}
Chain-of-Thought \cite{wei2022chain} refers to the general strategy of solving multi-step problems by decomposing them into a sequence of intermediate steps.
It has recently been applied extensively in a variety of problems such as mathematical reasoning \cite{ling2017program,cobbe2021training}, program execution \cite{reed2015neural,nye2021show}, commonsense or general reasoning \cite{rajani2019explain,clark2020transformers,liang2021explainable,wei2022chain}, and robotics \cite{xu2018neural,zhang2021hierarchical,jia2022learning,gu2022multi,yang2022chain,shridhar2023perceiver,james2022q}.
Similar ideas in the context of HRL can date back to Feudal RL \cite{dayan1992feudal} and the option framework \cite{sutton1999between}.
Inspired by these approaches, ours focuses on the imitation learning setup for low-level control tasks.
Note that while Procedure cloning \cite{yang2022chain} shares a similar name to our paper, it suffers from certain limitations that make it much less applicable (see Appendix \ref{app:more_dis}). 

\paragraph{Demonstrations for Robotics Tasks}
In practice, the optimality assumption of the demos is usually violated for robotics tasks. 
Demos involving low-level actions primarily come in three forms: human demo captured via teleoperation \cite{kumar2015mujoco,vuong2021single}, expert demo generated by RL agents \cite{mu2020refactoring,mu2021maniskill,chen2022system,jia2022improving}, or those found by planners (e.g., heuristics, sampling, search) \cite{gu2023maniskill2,qureshi2019motion,fishman2022motion}.
These demos are in general sub-optimal due to either human bias, imperfect RL agents, or the nature of the planners. 
In this paper, we largely work with demos generated by planners in ManiSkill2 \cite{gu2023maniskill2}, a benchmark not currently saturated for IL while being adequately challenging
(see details in Sec. \ref{sec:ms2}).

\section{Preliminaries}
\paragraph{MDP Formulation} 
One of the most common ways to formulate a sequential decision-making problem is via a Markov Decision Process, or MDP \cite{howard1960dynamic}, defined as a 6-tuple $\langle S, A, \mathcal{T}, \mathcal{R}, \rho_0, \gamma\rangle$, with a state space $S$, an action space $A$, a Markovian transition probability $\mathcal{T}: S \times A \rightarrow \Delta(S)$, a reward function $\mathcal{R}: S \times A \rightarrow \mathbb{R}$, an initial state distribution $\rho_0$, and a discount factor $\gamma \in[0,1]$. An agent interacts with the environment characterized by $\mathcal{T}$ and $\mathcal{R}$ according to a policy $\pi: S \rightarrow \Delta(A)$.
We denote a trajectory as $\tau_{\pi}$ as a sequence of $(s(0), a(0), s(1), a(1), ..., s(t), a(t))$ by taking actions according to a policy $\pi$.
At each time step, the agent receives a reward signal $r_t \sim \mathcal{R}(s(t), a(t))$.
The goal is to find the optimal policy $\pi^*$ that maximizes the expected return $\mathbb{E}_{\tau \sim \pi} [\sum_t \gamma^t r_t]$.
Remarkably, in robotics tasks and many real-world applications, the reward is at best only sparsely given (e.g., a binary success signal) or given only after the trajectory ends (non-Markovian). 

\paragraph{Behavior Cloning}
The most straightforward approach in Imitation Learning (IL) is Behavior Cloning (BC), which assumes access to pre-collected demos $D = \{(s(t), a(t))\}_{t=1}^N$ generated by expert policies and learns the optimal policy with direct supervision by minimizing the BC loss $\mathbb{E}_{(s,a)\sim D} [-\log \pi(a|s)]$.
It requires the learned policy to generalize to states unseen in the demos since at test time $\tau_{\pi}$ goes beyond the distribution from the demos, a challenge known as distribution shift \cite{ross2010efficient}. 
Recently, several methods, particularly those based on Transformers, relax the Markovian assumption, i.e.,
the policy is represented as $\pi(a(t)|s(t),...,s(t-T+1))$ or $\pi(a(t)|s(t),a(t),...,s(t-T+1),a(t-T+1))$ by considering a context history of size $T$.
This change was empirically shown to be advantageous.

\paragraph{Behavior Transformer} \label{sec:bet}
Behavior Transformer (BeT) \cite{shafiullah2022behavior} was proposed to tackle noisy (multi-modal) demonstrations for BC.
While relying on an architecture similar to the Decision Transformer (DT) \cite{chen2021decision}, it leverages a center plus offset representation of action signals, inspired by the object detection community.
Specifically, a (k-mean) clustering process is applied upfront to all actions in the demos to partition the action space into $K$ regions.
The model adopts one classifier and $K$ separate regressors (each for one region) to predict its actions.
This approach drastically improves optimization efficiency for BC.
During inference, BeT samples from the predicted region and its associated offset regression.
For simplicity, we always select the region with the largest predicted score (comparable performance). 
DT conditions action predictions based on a history of intertwined actions and states while BeT solely on states.
\textit{We adopted the DT's approach in our BeT variant (both as the BeT baseline and in CoTPC)} which we found to perform better because conditioning further on actions better distinguishes between noisy demos.

\begin{figure*}[ht]
\begin{center}
\includegraphics[width=0.9\linewidth]{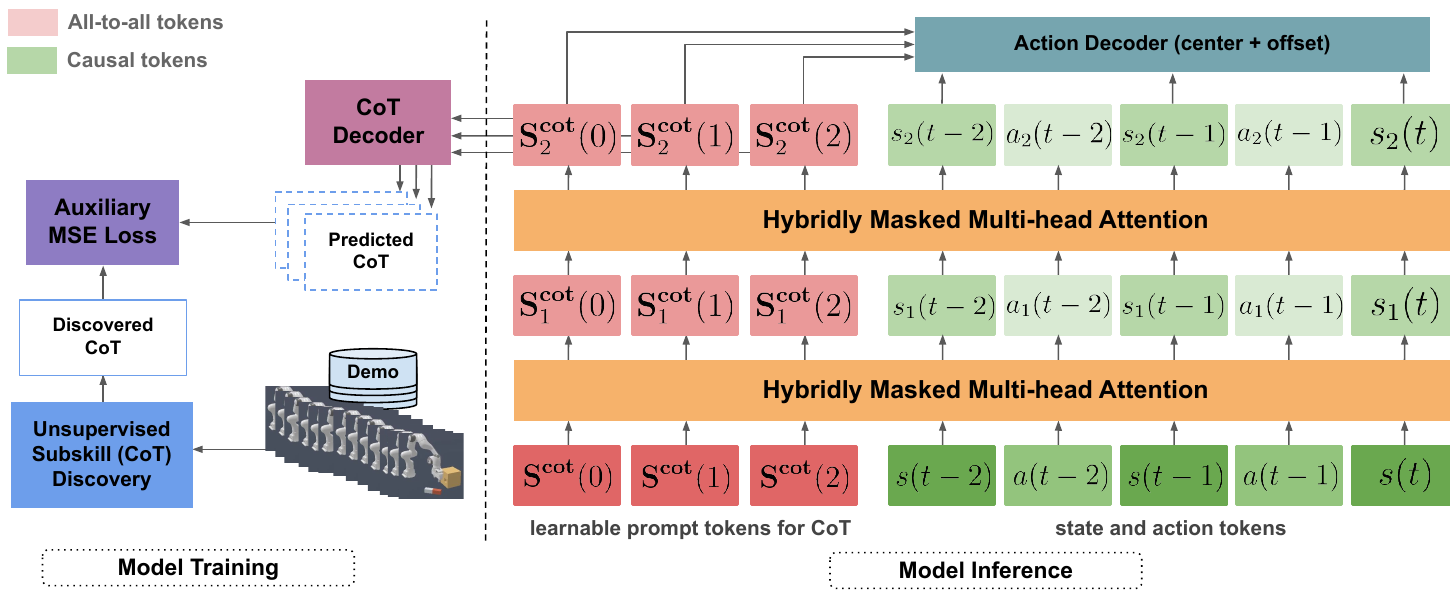}
\vskip -0.05in
\caption{During training, CoTPC learns to jointly predict (1) the next \& the last subskill from each CoT token and (2) the low-level actions from each CoT token (for action offset) and state token (for action center). See details in Sec. \ref{sec:coupled_action_and_cot}. During inference, when the CoT decoder is not used, the low-level actions are predicted under the guidance of the dynamically updated CoT features. The CoT tokens are all-to-all (can see any tokens). The state and action tokens are causal (can only see previous and CoT tokens). Only 2 attention layers and 3 timesteps are shown for better display.}
\label{fig:teaser}
\end{center}
\end{figure*}

\section{Subskill Sequence as Chain-of-Thought} \label{sec:rescue}
We propose two things.
(1) an observation space-agnostic strategy that efficiently discovers the subskill sequences from the demonstrations as CoT supervision in an unsupervised manner and (2) a novel Transformer-based design that effectively adopts the hierarchical principles, i.e., the subskill planning, in imitation learning. 
We name our approach Chain-of-Thought Predictive Control (\textbf{CoTPC}).

\subsection{Unsupervised Discovery of Subskills from Demos} \label{sec:key_state_selection}
We observe that many low-level control tasks naturally consist of sequences of subskills.
In a successful trajectory, there exist key observations, each of which marks the completion of a subskill.
For instance, in Moving Maze illustrated in Fig. \ref{fig:envs}, the two bridges naturally divide the task into three subskills.
We denote this multi-step subskill decomposition of a task into a chain of planning steps as its chain-of-thought (CoT), since the chain is considered a ``thought process'' for the task.
The CoT provides coherent and succinct behavior guidance - typical benefits of hierarchical policy learning.
Formally, for each trajectory $\tau \in D$, we define its CoT as a subsequence of its observations $F_{cot}(\tau) \triangleq \{s^{\textbf{cot}}(i)\}_{i=1}^{m}$, each of which marks the boundary of two adjacent subskill segments in the trajectory.
Here we have $i$ as the index of the $m$ subskills.

\begin{figure}
    \centering
    \includegraphics[width=1.0\linewidth]{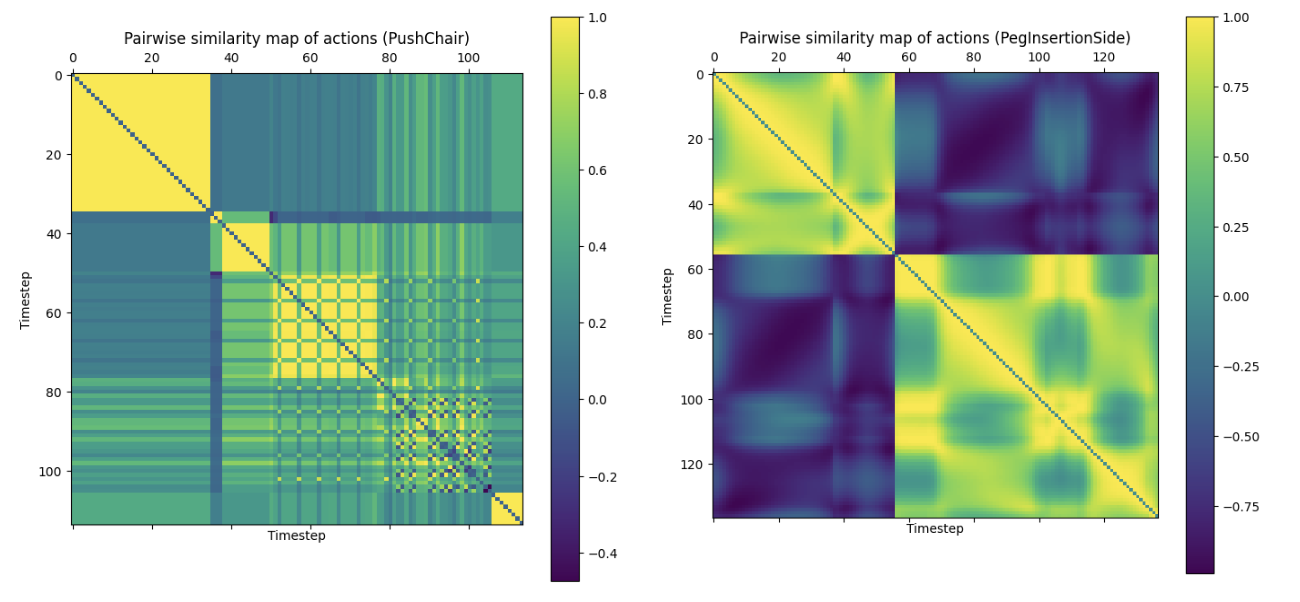}
 
    \vskip -0.2in
    \caption{Pairwise similarities of actions at different timesteps in two trajectories for Push Chair (\textbf{left}) and Peg Insertion (\textbf{right}). Action spaces are delta joint velocity and delta joint pose. Visually identifiable blocks along the diagonal are grouped, where actions are temporarily close and functionally similar. This corresponds very well with human intuition of subskills (see Appendix \ref{app:cos}). 
    }
    \label{fig:simmap}
    \vskip -0.3in
\end{figure}

How do we define subskill segments and how do we find them? 
Our base assumption is that actions within the same subskill are temporally close and functionally similar.
We, therefore, propose to \textit{group contiguous actions into segments} using a training-free procedure that finds the change points of the action sequences, which are modeled as time series.
Specifically, for each trajectory $\tau$, we consider its $p$-dimensional action at time $t$ as a vector $a(t) \in \mathcal{R}^{p}$. 
The subskill discovery process is formulated as an optimization problem to first find the change point indices $\{\tau_i\}_{i=1}^m$ of the action sequence in $\tau$ with
\begin{align*}
\tau_1, &\tau_2, ..., \tau_m = \argmin_{c_1 ... c_m} \mathcal{J}(\tau, \{c_i\})  \\
\mathcal{J}(\tau, \{c_i\}) &= \sum_{i=1}^m [\mathcal{C}(a(c_{i-1} + 1), …, a(c_i)) + \beta] \\
\mathcal{C}(v_1, …, &v_n) = \frac{1}{n} \sum_{i=1}^n d_{cos}(v_i, \bar{v}), \ \ \bar{v} = \frac{1}{n} \sum_{i=1}^n v_i 
\end{align*}
For convenience, we define $c_0 = -1$ as $\tau$ starts at $t=0$, and the last change point is defined at the last time index $|\tau|-1$ when the trajectory (and the last subskill) ends. 
We use $d_{cos}$, the Cosine distance (see discussion of this in Appendix \ref{app:metric}), and $\beta$ is a hyperparameter penalizing a large number of subskills $m$.
For instance, with $\beta$ = 0, every timestep becomes a change point, resulting in an over-segmentation of the demo trajectory. 
This optimization problem is typically solved by dynamic programming in a per-trajectory manner.
Here we adopt PELT \cite{Killick_2012}, an improved method with additional techniques such as pruning.
We provide a visualization of similarity maps across actions in the same trajectory in Fig. \ref{fig:simmap} to illustrate our intuitive action grouping strategy for subskill discovery.
Finally, we use the \textit{observations} $\{s^{\textbf{cot}}(i) \triangleq s(\tau_i)\}_{i=1}^{m}$ instead of the actions as the discovered CoT.

The number of subskills in a task is not known \textit{a priori}, since demonstrations of the same task can still have variability in terms of the starting configurations and execution difficulty.
For instance, in Push Chair, there is an absence of some subskills in the demos due to the varied initial conditions and the nature of the task. 
To achieve such flexibility, our approach only enforces $m$ as the maximum number of subskills to be predicted during inference and we assume that all demo trajectories in a task share the same subskill ``supersequence''.
In our experiments, we find $\beta$ relatively robust (see Appendix \ref{app:beta} for details).

We find our approach discovers meaningful subskills for a diverse set of tasks with \textit{different action spaces}, based on a qualitative evaluation of the key observations at the discovered change points.
For instance, the execution of Peg Insertion consists of reaching, grasping, aligning the peg with the hole, micro-adjusting, and steady insertion of the peg.
We illustrate some subskill discovery results in Appendix \ref{app:cos}. 
Besides being unsupervised, another major benefit of our approach is that it is observation space-agnostic with the discovery invariant of the specific sensor setup (e.g., camera angles, which are usually tuned ad-hoc for the specific task). 

\subsection{Subskill Guided Action Modeling} 
In this section, we introduce our Transformer-based design that learns to model the CoTs and the actions with the supervision acquired by the subskill discovery process. 

\subsubsection{Learnable Prompt Tokens for CoT with Hybrid Masking} 
We base our architecture on a modified version of the recently proposed Behavior Transformer (BeT) \cite{shafiullah2022behavior}, which empowers GPT \cite{brown2020language} with a discrete center plus continuous offsets prediction strategy for modeling diverse and noisy action sequences.
Please refer to Sec. \ref{sec:bet} for a brief introduction to the BeT architecture.
To predict CoTs, we propose to add a set of learnable prompt tokens \cite{zhou2022learning}, denoted CoT tokens, at the beginning of the state and action context history.
We train these tokens to extract features from the sequence to predict CoTs together with a CoT decoder (similar to the object query tokens in Detection Transformer \cite{carion2020end}).
We design a \textit{hybrid} masking regime, where during inference, the CoT tokens are all-to-all and can observe all action and state tokens in the context history, and the state or action tokens attend to those in the past (standard causal mask) including the CoT tokens.
In this way, the action decoding is guided by the extracted CoT features corresponding to the CoT tokens.

Formally, given a context size $T$, we choose to use $T$ CoT tokens ${\mathbf{S}^{\textbf{cot}}(i)}$, each of which is a different learnable prompt token $\in \mathcal{R}^{128}$, where $128$ is the embedding dim of the Transformer.
Accordingly, the demo segment up to time $t$ consists of $T$ CoT tokens, $T$ state and $T-1$ action tokens:
\begin{align} \label{eq:traj}
\tau_T(t) &= \mathbf{S}^{\textbf{cot}}(0),\ ...,\ \mathbf{S}^{\textbf{cot}}(T-1),\ s(t-(T-1)), \nonumber \\ 
a(&t-(T-1)),\ ...,\ s(t-1),\ a(t-1),\ s(t) 
\end{align}
By applying hybridly masked multi-head attention $\mathcal{M}[\cdot]$. the features from a total of $J$ attention layers are represented as 
\begin{align*}
h_j( \tau_T (t) ) &= \mathcal{M}[F_{enc}(\tau_T (t))], \ \ j = 1 \\
h_j( \tau_T (t)) &= \mathcal{M}[h_{j-1}( \tau_T (t))], \ \ j > 1
\end{align*}
where $F_{enc}$ encodes each action token and state token by encoder $f_a(\cdot)$ and $f_s(\cdot)$, respectively (no encoder for the CoT tokens). Here we omit the position embeddings and the additional operations between the attention layers as in standard Transformers.
As a result, features from the last attention layer can be written as
\begin{align*}
h_J( \tau_T&(t) ) = \mathbf{S}_J^{\textbf{cot}}(0),\ ...,\ \mathbf{S}_J^{\textbf{cot}}(T-1),\ s_J(t-(T-1)), \\
&a_J(t-(T-1)),\ ...,\ s_J(t-1),\ a_J(t-1),\ s_J(t)
\end{align*}
The CoT features from CoT tokens are then fed into the CoT decoder $g_{cot}$ to predict $\{g_{cot}(\mathbf{S}_J^{\textbf{cot}}(t))\}^{T-1}_{t=0}$ (see why $T$ CoT tokens and $T$ outputs shortly).
In our experiment, we use a 2-layer MLP with ReLU activation for $g_{cot}$.

While the learnable prompt tokens in Transformers are widely used in tasks regarding language, vision \& language, etc., it is \textit{under-explored when applied to low-level continuous control problems}, which is highly non-trivial as demonstrated in our ablation studies.


\subsubsection{Coupled Action and Subskill Predictions with Shared CoT Tokens}
\label{sec:coupled_action_and_cot}

Inspired by BeT, we adopt a pair of center and offset decoders for predicting actions, denoted as $g^{\textbf{ctr}}_a(\cdot)$ and $g^{\textbf{off}}_a(\cdot)$.
For an action $a(t')$, we use $g^{\textbf{ctr}}_a(\cdot)$ to predict its center as
\[
g^{\textbf{ctr}}_a(s_J(t')),\ \ t' \in \{t-(T-1), ..., t-1, t\}
\]
where $t$ follow the previous notation in $h_J( \tau_T(t) )$.
Contrary to the original BeT, we predict its offset with 
\[
g^{\textbf{off}}_a(\mathbf{S}_J^{\textbf{cot}}(t'')),\ t'' = t' - [t - (T-1)]
\]
We essentially use features from the $i^{th}$ CoT token to predict the $i^{th}$ action's offset component.
Accordingly, we need $T$ CoT tokens so that each CoT token's features are used to predict the action offset at one of the $T$ different timesteps.
See Fig. \ref{fig:teaser} for the overall architecture.
As the CoT decoder $g_{cot}(\cdot)$ uses the same inputs as the action offset decoder, this couples action and CoT predictions for stronger subskill guidance.
An alternative design is to use only one CoT token to produce CoT predictions as $g_{cot}(\textbf{S}^{\textbf{cot}})$.
The actions offset prediction is based on the state tokens instead (similar to vanilla BeT).
However, we find this decoupled strategy produces policies much less generalizable.
An even more coupled prediction approach is to force the center decoder $g^{\textbf{ctr}}_a(\cdot)$ to also share the inputs with $g_{cot}(\cdot)$, which leads to optimization instabilities since CoT tokens hold direct responsibilities for all predictions.
Please see our ablation studies for empirical evidence.

As for subskill prediction, we formulate an autoregressive prediction strategy by training $g_{cot}(\cdot)$ to decode both the next subskill and the very last one from features of every CoT token.
At time $t$ for a trajectory $\tau$, the next subskill, denoted $s^{\textbf{cot}}_{next}(t) = s(\tau_i)$, is one with the first $\tau_i$ such that $\tau_i > t$.
The last subskill is always the end of the demo trajectory where the task is completed, denoted $s^{\textbf{cot}}_{last}(t) = s(|\tau|-1)$.
Formally, assuming the $q$-dimensional state or observation space $\mathcal{R}^q$, for each CoT token $\textbf{S}^{\textbf{cot}}(t)$ with $t=\{0, 1, ..., T-1\}$, the CoT decoder produces $g_{cot}(\mathbf{S}_J^{\textbf{cot}}(t)) \in \mathcal{R}^{2q}$.
The $T$ CoT predictions make each CoT token extract useful features for subskill planning which benefit action offset predictions at all timesteps. 
We empirically verified that this strategy of performing both immediate and long-term planning outperforms the two alternatives.

Due to the autoregressive nature of CoT predictions, our approach admits a \textit{flexible} number of subskills across different demo trajectories of the same task.
Moreover, as the CoT features are updated dynamically at each timestep, CoTPC can handle tasks involving \textit{dynamic} controls (e.g., Moving Maze and Push Chair).
We name our approach after CoT since our subskill predictions resonate with CoTs in LLMs, where the text outputs are generated jointly with the reasoning chain.
Note that, during inference, the CoT decoder is not used and only the CoT features are (for predicting action offsets).
See Sec. \ref{sec:limit} for a further discussion on the topic.
Also see Fig. \ref{fig:abl} in Appendix \ref{app:dataflow} for a comparison of data flows of different versions regarding the action decoders (both offset and center) and the CoT tokens.

\subsubsection{Model Training}
The overall pipeline is illustrated in Fig. \ref{fig:teaser}. 
The model is trained with behavior cloning loss as well as the auxiliary CoT prediction loss $\mathcal{L}_{cot}$ based on MSE (weighted by a coefficient $\lambda = 0.1$), yielding the overall objective as

\begin{align*}
\mathcal{L}_{total} &= \mathop{\mathbb{E}}_{(s_t, a_t)\in D}\mathcal{L}_{bc}(a_t, \hat{a}_t) \ + \\ \mathop{\mathbb{E}}_{\tau_T(t) \sim D}\ &\mathop{\mathbb{E}}_{t'}\ \mathcal{L}_{cot}([s^{\textbf{cot}}_{next}(t'), s^{\textbf{cot}}_{last}(t')],g_{cot}(\textbf{S}^{\textbf{cot}}_J(t'')))
\end{align*}
where $\hat{a}_t$ is the predicted action via $g^{\textbf{ctr}}_a(\cdot)$ and $g^{\textbf{off}}_a(\cdot)$. 
Following the notation from Eq. \ref{eq:traj} and the previous section, $\tau_T(t)$ is a randomly sampled segment of context length $T$ from the demonstration set $D$, and $t'' = t' - [t - (T-1)]$ is the shifted index for the CoT tokens.
There are $T$ loss terms for CoT since there are $T$ CoT tokens.



During training, we apply random attention masks to the action and state tokens so that the CoT tokens attend to a context of varied lengths. 
For each data point in the mini-batch, we uniformly sample $l \in \{0, 1, ..., T-1\}$ and manually set the attention mask from all the CoT tokens to the state and action tokens corresponding to $s(t’)$ and $a(t’)$ to zero (i.e., disabled) if $t’ > t - l$. 
The intuition of this regularization technique is that we should predict similar future subskills given segments of different lengths of the same history. 
Moreover, since CoT tokens are all-to-all, without such a technique, they might attend to future tokens during training to make subskill predictions trivial to some extent.
See relevant ablation studies in Sec. \ref{sec:abl}.

\begin{figure*}[ht]
\begin{center}
\centerline{\includegraphics[width=0.93\linewidth]{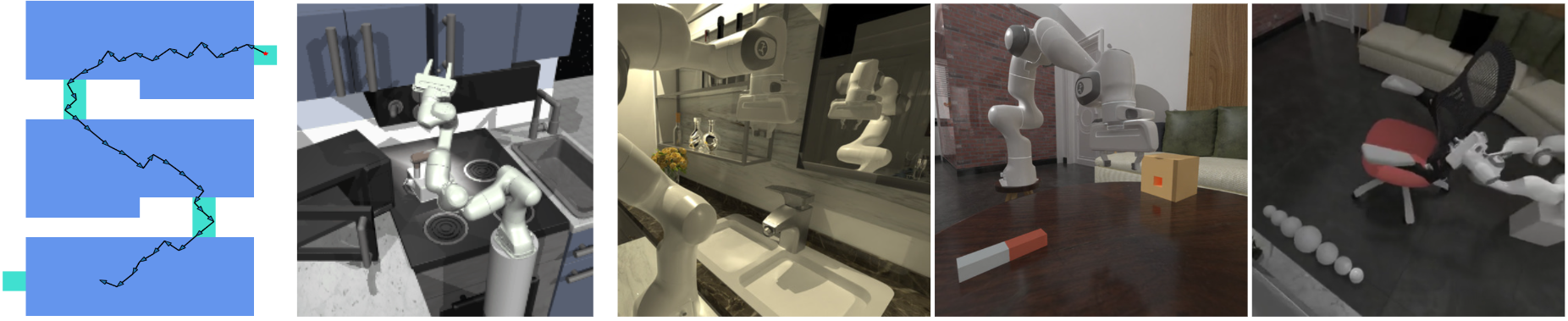}}
\caption{Illustration of the Moving Maze (\textbf{left}), Franka-Kitchen (\textbf{middle}) and some sampled tasks from ManiSkill2 (\textbf{right}), namely Turn Faucet, Peg Insertion, and Push Chair. See detailed descriptions in Sec. \ref{sec:maze}, \ref{sec:kitchen} and \ref{sec:ms2}, respectively.}
\label{fig:envs}
\end{center}
\vskip -0.1in
\end{figure*}

\begin{figure}[ht]
\vskip 0.05in
\begin{center}
\centerline{\includegraphics[width=\linewidth]{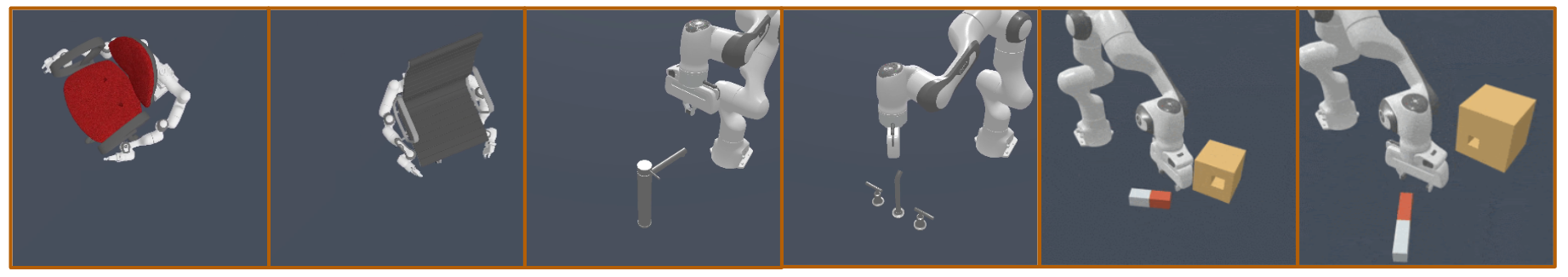}}
\caption{Sampled geometric variations for Push Chair, Turn Faucet, and Peg Insertion. The sizes of peg \& box and the relative locations of the hole vary across different env. configs.}
\label{fig:variations}
\end{center}
\vskip -0.2in
\end{figure}

\section{Experiments} \label{sec:experiments}
In this section, we present our main experimental results as well as ablation studies. 
While existing benchmarks are mostly saturated for IL (DMControl \cite{tassa2018deepmind}, D4RL \cite{fu2020d4rl}, etc.) or lack demo data (e.g., MineDojo \cite{fan2022minedojo}), we choose a diverse range of tasks that lie in between.
We first examine our approach using a 2D continuous-space Moving Maze and a variant of Franka-Kitchen \cite{gupta2019relay}.
We then perform extensive comparisons with 5 object manipulation tasks (ranging from relatively easy to very challenging) from ManiSkill2 \cite{gu2023maniskill2}.
These tasks are of several categories (navigation, static/mobile manipulation, and soft-body manipulation, etc.), various action spaces (delta joint pose, delta velocity, etc.), with different sources of the demos (human, heuristics, etc.), and pose distinctive challenges (see Sec. \ref{sec:ms2} \& Appendix \ref{app:challenges} for details).

\subsection{Moving Maze} \label{sec:maze}
We present a 2D maze with a continuous action space of displacement $(\delta x, \delta y)$. 
As shown in Fig. \ref{fig:envs} (left), in this s-shaped maze, 
the agent starts from a location randomly initialized inside the top right square region (in green). 
The goal is to reach the bottom left one (also in green).
Upon each environment reset, the two regions and the two rectangular bridges (in green) have their positions randomized.
Each of them except for the top square moves (independently) back and forth with a randomized constant speed.
Once the agent lands on a moving block, the block becomes static.
The agent cannot cross the borders of the maze (but will not die from doing so).
For simplicity, we adopt a 10-dim state observation consisting of the current location of the agent and the four green regions. 
This task requires \textit{dynamic control/planning}.

\paragraph{Demonstrations} To enable policy learning from demonstrations, we curate demo trajectories (each with a different randomized environment configuration) 
by adopting a mixture of heuristics and an RRT-style planner with hindsight knowledge not available at test time (following recent work \cite{gu2023maniskill2} for leveraging machine-generated demos).

\paragraph{Training and Evaluation}
For this task, we compare CoTPC with vanilla BC, 
Behavior Transformer (BeT) \cite{shafiullah2022behavior}, Decision Transformer (DT) \cite{chen2021decision}.
DT was originally proposed for offline RL with demonstrations of dense rewards. 
We adapt it for the BC setup by ignoring the reward tokens. 
We add action tokens to BeT (like in DT) and build CoTPC on top of BeT. 
We implement CoTPC, DT, and BeT with the shared Transformer configuration for a fair comparison.
We train all methods on 400 demo trajectories of different env. configs and evaluate on 100 unseen ones (results in Tab. \ref{tab:maze_and_kitchen}).

\subsection{Franka Kitchen} \label{sec:kitchen}
Different variants and task setups of the Franka Kitchen environment have been studied previously \cite{gupta2019relay,sasaki2020behavioral,fu2020d4rl}.
We propose a setting where the agent is asked to complete 4 object manipulations (out of 7 different options) in an order specified by the goal.
We use a strict criterion, i.e., the task succeeds when all 4 sub-tasks are completed.
The sub-tasks need to be done in the requested order to succeed.
The environment will terminate when a sub-task other than the specified 4 is performed.
The action space is based on the joint velocity (8-dim) of the robot.
We use the original state observation appended with the modified goal embedding.  

\paragraph{Demonstrations}
We replay a subset of the human demonstrations originally proposed in \cite{gupta2019relay}.
Specifically, we use 50 demo trajectories of length ranging from $150$ to $300$ and relabel them with what sub-tasks are performed and in what order, for each of the trajectories.
As a result, many ordered sub-task combinations admit at most one demo trajectory.
See more details in the Appendix. 

\paragraph{Training and Evaluation}
We use the same set of baselines as in Moving Maze.
We evaluate using 90 unseen env. configs, which vary in initial scene configs (all ordered sub-task combinations have been observed in the demo, though).
This task requires generalizable IL due to the limited amount of human demos and the diverse set of ordered sub-task sequences.
Also see results in Tab. \ref{tab:maze_and_kitchen}.

\begin{table}[t]
\caption{Test performance on Moving Maze and (a variant of) Franka Kitchen. SR (\%) is the task success rate (for Franka Kitchen it means completion of all 4 sub-tasks). \# s-tasks means the avg. number of completed sub-tasks per trajectory rollout. The best results are \textbf{bolded}.}
\label{tab:maze_and_kitchen}
\vskip -0.1in
\begin{center}
\begin{scriptsize}
\begin{tabular}{l c c c c}
\toprule
& Vanilla BC & DT & BeT & CoTPC (ours) \\
\cmidrule(lr){2-5}
Maze (SR) & 9.0 & 23.0 & 33.0 & \textbf{44.0} \\
Kitchen (\#s-tasks/SR) & 1.7/6.7 & 1.6/6.7 & 1.8/14.4 & \textbf{2.1}/\textbf{25.6}  \\
\bottomrule
\end{tabular}
\end{scriptsize}
\end{center}
\vskip -0.1in
\end{table}

\begin{table*}[t]
\caption{Test performance (unseen and 0-shot success rate) for ManiSkill2 tasks with state observations. The best results are \textbf{bolded}. Diffusion-based methods are generally slower and might be less data-efficient (see a discussion \href{https://drive.google.com/file/d/1YUPi2quM20viKrntyt01dL_sArynVPak/view}{here}), making dynamic control challenging. See additional mean \& std. of the results among 3 runs in Appendix \ref{app:var}.}
\vskip 0.1in
\label{tab:more_main_results}
\begin{center}
\begin{small}
\begin{sc}
\begin{tabular}{l c c cc cc}
\toprule
  & \multicolumn{1}{c}{Stack} & \multicolumn{1}{c}{Peg}  & \multicolumn{2}{c}{Turn} & \multicolumn{2}{c}{Push} \\
  & \multicolumn{1}{c}{Cube} & \multicolumn{1}{c}{Insertion} & \multicolumn{2}{c}{Faucet} & \multicolumn{2}{c}{Chair}  \\
\midrule
 & {\scriptsize unseen} & {\scriptsize 0-shot} & {\scriptsize unseen} & {\scriptsize 0-shot} & {\scriptsize unseen} & {\scriptsize 0-shot} \\
\cmidrule(lr){2-2}\cmidrule(lr){3-3}\cmidrule(lr){4-5}\cmidrule(lr){6-7}
Vanilla BC & 1.0 & 0.0 & 0.0 & 0.0 & 0.0 & 0.0 \\
Decision Transformer & 19.0 & 17.5 & 40.0 & 27.0 & 25.6 & 17.0 \\
Decision Diffuser & 26.0 & 12.6 & 17.0 & 5.0 & \textbf{56.0} & 20.0 \\
Diffusion Policy & 84.5 & 57.6 & \textbf{52.0} & 35.0 & 54.0 & 38.0 \\
Behavior Transformer & 73.0 & 42.5 & 49.6 & 32.5 & 44.0 & 33.4 \\
CoTPC (ours) & \textbf{86.0} & \textbf{59.3} & 50.0 & \textbf{39.3} & 51.2 & \textbf{41.0} \\
\bottomrule
\end{tabular}
\end{sc}
\end{small}
\end{center}
\end{table*}
\vskip -0.05in

\begin{table*}[t]
\caption{Test performance (success rate) on the unseen and the 0-shot setup for ManiSkill2 tasks for point cloud observations. Pour only supports visual observations.  The best results are \textbf{bolded}. We only show the closely related BeT baseline here.}
\vskip 0.1in
\label{tab:pcd_results}
\begin{center}
\begin{small}
\begin{sc}
\begin{tabular}{l c c c cc cc}
\toprule
  & \multicolumn{1}{c}{Cube} & \multicolumn{1}{c}{Peg}  & \multicolumn{1}{c}{Pour} & \multicolumn{2}{c}{Faucet} & \multicolumn{2}{c}{Chair} \\
\midrule
 & {\scriptsize unseen} & {\scriptsize 0-shot} & {\scriptsize unseen} &  {\scriptsize unseen} & {\scriptsize 0-shot} & {\scriptsize unseen} & {\scriptsize 0-shot} \\
\cmidrule(lr){2-2}\cmidrule(lr){3-3}\cmidrule(lr){4-4}\cmidrule(lr){5-6}\cmidrule(lr){7-8}
Behavior Transformer & 70.0 & 35.0 & 24.0 & 50.0 & 20.0 & 26.0 & 13.4 \\
CoTPC (ours) & \textbf{81.0} & \textbf{44.0} & \textbf{32.0} & \textbf{58.0} & \textbf{27.5} & \textbf{32.0} & \textbf{16.7} \\
\bottomrule
\end{tabular}
\end{sc}
\end{small}
\end{center}
\end{table*}
\vskip -0.05in

\subsection{ManiSkill2} \label{sec:ms2}
ManiSkill2 features a variety of low-level object manipulation tasks in environments with realistic physical simulation (e.g., fully dynamic grasping motions).
We choose 5 tasks (see Fig. \ref{fig:envs}).
Namely, Stack Cube for picking up a cube, placing it on top of another, and the gripper leaving the stack; 
Turn Faucet for turning on different faucets; 
Peg Insertion for inserting a cuboid-shaped peg \textit{sideways} into a hole in a box of different geometries and sizes;
Push Chair for pushing different chair models into a specified goal location (via a mobile robot);
and Pour for pouring liquid from a bottle into the target beaker with a specified liquid level.
Push Chair adopts a delta joint velocity control (19-dim, dual arms with mobile base); Pour adopts delta end effector pose control (8-dim); the rest uses delta joint pose control (8-dim).
We perform experiments with both state and point cloud observations.

\paragraph{Task Complexity}
The challenges of these tasks come from several aspects.
Firstly, all tasks have all object poses fully randomized (displacement around 0.3m and $360\degree$ rotation) upon environment reset (this is in contrast to environments such as Franka Kitchen). 
Secondly, Turn Faucet, Peg Insertion, and Push Chair all have large variations in the geometries and sizes of the target objects (see illustrations in Fig. \ref{fig:variations}).
Moreover, the faucets are mostly pushed rather than grasped during manipulation (under-actuated control), the holes have 3mm clearance (requiring high-precision control) and it needs at least half of the peg to be pushed sideway into the holes (harder than similar tasks in other benchmarks \cite{xu2019compare}), the chair models are fully articulated with lots of joints, and the pouring task requires smooth manipulation without spilling the liquid.
Moreover, ManiSkill2 adopts impedance controllers that admit smoother paths (important for tasks like Pour) than the position-based ones while at the cost of harder low-level action modeling (e.g., actuators can be quite laggy).

\paragraph{Demonstrations}
The complexity of the tasks also lies in the sub-optimality of the demos (e.g., vanilla BC struggles on all 5 tasks).
The demos are generated by multi-stage motion-planing and heuristics-based policies (with the help of privileged information in simulators).
We use $500$ demo trajectories for Stack Cube and Turn Faucet (distributed over 10 faucets), $1000$ demos for Peg Insertion and Push Chair (distributed over 5 chairs), and $150$ demos for Pour. 

\paragraph{Training and Evaluation}
Besides vanilla BC, DT, and BeT, we add Decision Diffuser (DD) \cite{ajay2022conditional} and Diffusion Policy (DP) \cite{chi2023diffusion} as two baselines, which explore the diffusion model \cite{ho2020denoising} for policy learning by either first generating a state-only trajectory and then predicting actions with an acquired inverse dynamics model or directly performing diffusion steps to predict actions.
Remarkably, diffusion-based policies rely on repeated denoising steps and are \textit{generally much slower than the likes of BeT or CoTPC, making dynamic control challenging}.
The tasks in ManiSkill2 feature diverse object-level variations and provide good insights into both how effective an imitator can learn the underlying behavior and how generalizable it is.
We evaluate using the 5 tasks in both unseen (seen objects but unseen scene configs) and 0-shot (unseen object geometries) setup.
Specifically, we have all but Peg Insertion with the unseen setup and Turn Faucet, Push Chair \& Peg Insertion with the 0-shot setup.
We use task success rate (SR) as the metric.
Results are reported in Tab. \ref{tab:more_main_results} for state observation and Tab. \ref{tab:pcd_results} for point cloud observations, where we demonstrate the clear advantages of CoTPC.
To avoid tuning point cloud encoders (we use a lightweight PointNet \cite{qi2017pointnet} trained from scratch) for each baseline, we only use the BeT baseline for point cloud observations.
For the main results of all methods on ManiSkill2, we report the best performance among 3 training runs over the last 20 checkpoints, an eval protocol adopted by \cite{chi2023diffusion}.
See additional mean \& std. results among 3 runs in Appendix \ref{app:var}.

\begin{table*}[t]
\caption{Results from the ablation studies (unseen SR for Push Chair and 0-shot SR for Peg Insertion).}
\label{tab:abl}
\begin{center}
\begin{scriptsize}
\begin{sc}
\begin{tabular}{l c c c c c c c c c c}
\toprule
& decoupled & only last & only next & random & vanilla & o-shared & swapped & BeT+prompt & -rand. mask & CoTPC \\
\cmidrule(lr){2-2}\cmidrule(lr){3-5}\cmidrule(lr){6-8}\cmidrule{9-10}
Peg & 47.0 & 52.0 & 49.0 & 41.0 & 45.0 & 39.0 & 46.0 & 43.0 & 48.5 & 59.3\\
Chair & 36.0 & 36.0 & 37.0 & 31.0 & 35.0 & 29.0 & 32.0 & 32.7 & 34.2 & 41.0 \\
\bottomrule
\end{tabular}
\end{sc}
\end{scriptsize}
\end{center}
\vskip -0.1in
\end{table*}

\subsection{Ablation Studies} \label{sec:abl}
We present ablation studies on two tasks (Peg Insertion and Push Chair) from ManiSkill2 with state observations.
We summarize the results in Tab. \ref{tab:abl} and introduce the details in the rest of this section. 

\paragraph{Decoupled prediction of subskills and actions} 
In this variant denoted \texttt{decoupled}, we train another Transformer (denoted CoT Transformer) with the same state and action sequence as inputs to predict the CoT (the next and the last subskill) instead of relying on one Transformer to predict both actions and subskills. During inference, the subskills predicted by the CoT Transformer are fed into the original Transformer for predicting the actions. This decoupled subskill and action prediction strategy is inferior to the coupled one since the latter not only learns to leverage predicted subskills as guidance but also encourages better feature representation of the trajectory by sharing the features for these prediction tasks.
In our early study, we found the coupled strategy worked better among different action spaces, model sizes, and context lengths.

\paragraph{What subskills to predict from the CoT tokens?}
In the variant named \texttt{only last}, we ask the CoT decoder $g_{cot}(\cdot)$ to only predict the last subskill.
In the variant named \texttt{only next}, we ask it to only predict the next subskill.
We find that predicting both works the best as it provides the model with both immediate and long-term planning.
In the variant named \texttt{random}, we ask it to predict a single randomly selected observation from the future as a random subskill (not the ones from our subskill discovery process).
This leads to the worst results as the action guidance does not help much and can even be misleading.

\paragraph{Shared CoT tokens for subskill and action predictions}
A vanilla design of jointly predicting subskills and actions is to only use one CoT token with the center and offset decoders on top of the state tokens and the CoT decoder on top of the CoT token, denoted as \texttt{vanilla}.
We further force both the center and the offset decoders to take CoT tokens as inputs, denoted \texttt{o-shared}. This variant overly shares the CoT tokens for all decoders and suffers from optimization instabilities.
In the last variant, denoted \texttt{swapped}, we let the offset decoder take the action tokens as inputs and the center decoder take the CoT tokens.
We find this alternative setup performs worse.
We illustrate the data flow of these variants in Appendix \ref{app:dataflow}.

\paragraph{Does the gain come from extra network capacity?}
In the variant denoted \texttt{BeT+prompt}, we add additional prompt tokens to the BeT baseline (no auxiliary loss and no CoT decoder). This variant has the same capacity as CoTPC yet performs very similarly to the BeT baseline, indicating that the performance gain achieved by our approach does not solely come from the extra network capacity.

\paragraph{How helpful is random attention masking?}
We propose to randomly mask the attention between CoT tokens and action/state tokens during training. 
This technique is very helpful as the performance drops significantly without it (see variant \texttt{-rand.} \texttt{mask}).
This technique cannot be applied to the BeT baseline (which has no CoT token, and all other tokens in BeT already have attention masks manually set to 0 regarding any future tokens).
Note that the variant \texttt{BeT+prompt} applies this technique but trains without the auxiliary CoT loss.
Its inferior results indicate that this regularization technique alone cannot provide a meaningful performance boost for BeT either.

\subsection{Additional Experiments} \label{sec:real}
We perform two experiments to explore the possibilities for sim-to-real transfer of state observation-based CoTPC.
We present promising preliminary results in Appendix \ref{app:s2r}.

\section{Discussion and Limitation}
\label{sec:limit}

\paragraph{Observation space-agnostic subskill discovery} 
Our proposed subskill discovery only depends on actions (and works well for different action spaces) which provide good functional abstraction.
Utilizing additional observation, on the contrary, has major disadvantages. 
Firstly, visual observations are more high-dimensional than actions and more challenging. 
Secondly, using visual information depends on the specific sensor setup (camera angles, etc.), leading to less robust results. 
Alternatively, one can use visual features extracted from some pre-trained encoders, an approach that remains an open question \cite{hansen2022pre}. 
On the other hand, the human-designed state observation 
are not generally accessible even for tasks in simulators (e.g., Pour only supports visual observations with soft-body). 

\paragraph{Limitations} One limitation of CoTPC lies in the assumption that similar and nearby actions should be grouped into the same subskill. This can be violated with complex manipulations with many micro-adjustments (e.g., standing and balancing on a rope). In this case, extra force feedback might be required to distinguish subskills that are only subtly different.
Another limitation is that our method is purely offline and not invincible to sub-optimal demos.
We believe additional online fine-tuning makes it more robust.

\paragraph{Low-level vs. high-level tasks} 
Many existing work dealing with ``long-horizon'' robotic tasks (SayCan \cite{ahn2022can}, ALFRED \cite{shridhar2020alfred}, etc.) assumes that low-level control is solved or that the task hierarchy is given.
On the contrary, in this paper, we study better ways to learn to solve low-level control tasks with unsupervisedly discovered hierarchical information as supervisions.
We believe that CoTPC can be extended in a multi-task learning setup and given enough demonstration, used as a foundation model for low-level control.

\paragraph{Is CoTPC truly hierarchical?} While CoTPC is not strictly a model-based planning method and not “100\%” hierarchical since the CoT decoder is not used during inference, it implicitly uses the hierarchical information (CoT features) that is dynamically updated. 
The CoT decoder is a lightweight network and the CoT features are not significantly different from the predicted subskills.
We found that using a more expressive CoT decoder makes CoT features less predictive of subskills and thus harms action decoding.

\section{Conclusion}
In this work, we propose CoTPC, a hierarchical imitation learning algorithm for learning generalizable policies from scalable but sub-optimal demos.
We formulate the hierarchical principles in the form of chain-of-thought (i.e., subskill) guidance in offline policy learning with a novel Transformer-based design and provide an effective way to obtain subskill supervision from demonstrations in an unsupervised manner.  
We demonstrate that CoTPC can solve a wide range of challenging low-level control tasks, consistently outperforming many existing methods.

\section*{Impact Statement}
This paper presents work whose goal is to advance the field of learning-based control policies. 
There are many potential societal consequences of our work, none of which we feel must be specifically highlighted here.

\bibliography{example_paper}
\bibliographystyle{icml2024}

\newpage
\appendix
\onecolumn

\section{General Challenges of Imitation Learning from Sub-optimal Demonstrations for Low-level Control Tasks} \label{app:challenges1}

\paragraph{Non-Markovity}
While each trajectory in the demos can be represented by a Markovian policy, the Markovian policy linearly combined from them by perfectly imitating the combined demos can suffer from a negative synergic effect if there are conflicts across demos.
This is because the demos might be generated by different agents or different runs of the same algorithm.
It becomes even worse when the demonstrations themselves are generated by non-Markovian agents (e.g., human or planning-based algorithms).
Instead, a non-Markovian policy is more universal and can resolve conflicts by including history as an additional context to distinguish between different demos.  

\paragraph{Noisiness}
Sometimes the demo trajectories are intrinsically noisy with divergent actions produced given the same states.
For instance, a search-based planner returns more than one possible action given the same action and state history to solve the task.
At times, the demo actions are even distributed uniformly (e.g., with motion planning algorithms as demonstrators).
This leads to increased uncertainty and variance of the cloned policies and so higher compounding errors.
Note that multi-modality is a related but orthogonal issue \cite{shafiullah2022behavior}, i.e., when an unimodal estimate of the (continuous) action distribution leads to a significantly worse return.

\paragraph{Discontinuity}
For low-level control tasks, demo policies often consist of sharp value changes or topology changes (e.g., due to contact changes).
Such discontinuity in the underlying state-to-action mapping leads to difficulties in learning a robust and accurate model, thus harming generalizability.
A recent method \cite{florence2022implicit} deals with this by an energy-based implicit model in place of an explicit one.
While theoretically sound, it is shown \cite{shafiullah2022behavior} to be less practical for non-Markovian implicit models, and several later non-Markovian explicit models outperform it. 

\paragraph{Randomness}
The actual or apparent unpredictability usually exists in sub-optimal demonstrations either because the intermediate computations of the demonstrators are not revealed in the demos (e.g., the shortest paths generated by BFS do not reveal the intermediate search process), or the demonstrators are inherently non-deterministic (e.g., relying on rejection sampling).
Such a trait makes IL less robust as the decision-making patterns from demos might be unclear, hard to learn, and not generalizable \cite{paster2022you}.
For instance, in a continuous action space maze, a solution found by random search is more-or-less a winning lottery ticket, whose pattern might not be very generalizable.


\section{Specific Challenges of Tasks In our Evaluation for Behavior Cloning} \label{app:challenges}

There are multiple challenges presented in the tasks used in our evaluation. 
Firstly, it involves large geometric variations of articulated objects, which is very non-trivial and is under-explored in the literature (see \url{https://maniskill2.github.io/} for more details). 
Secondly, the dynamics are challenging regarding (1) high-precision manipulations (some ManiSkill2 tasks such as Peg Insertion entail a precision level of millimeters), (2) under-actuated control (e.g., Push Chair and Turn Faucet), and (3) harder low-level action modeling due to the underlying impedance controller.
Thirdly, the demonstrations are inherently sub-optimal due to either the underlying contact dynamics or how they are produced (noisy, discontinuous, etc.).
Finally, the demonstrations can be limited in quantity (e.g., in our setup of Franka Kitchen).

\section{Illustration of the Ablation Study on Network Data Flow} \label{app:dataflow}
\begin{figure}
    \includegraphics[width=0.95\linewidth]{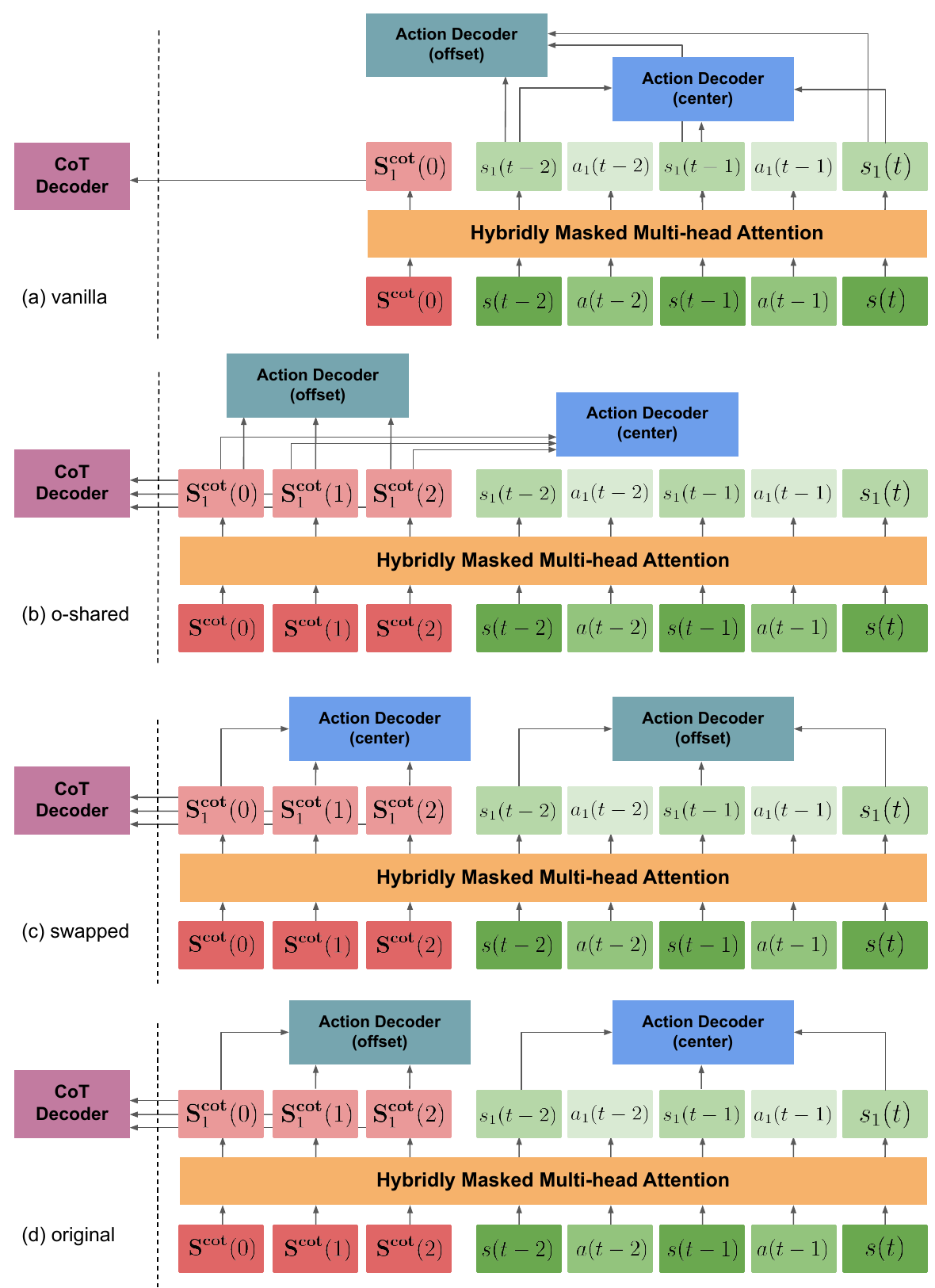}
    \caption{Illustration of the network data flow of our third ablation study for the variant named \texttt{vanilla}, \texttt{o-shared}, and \texttt{swapped} as well as the original design.}
    \label{fig:abl}
\end{figure}

To better explain the difference among the variants in our third ablation study in Sec. \ref{sec:abl} (i.e., regarding ``shared CoT tokens for subskill and action predictions''), we illustrate the network data flow of the three variants as well as the original design in Fig. \ref{fig:abl}.

\section{Additional Discussions of Related Work}
\label{app:more_dis}

\paragraph{Procedure Cloning (PC)}
PC \cite{yang2022chain} was recently proposed to use intermediate computation outputs of demonstrators as additional supervision for improving the generalization of BC policies.
However, it assumes full knowledge of the demonstrators, including the usually hidden computations that consist of potentially large amounts of intermediate results. 
For instance, in the graph search example used in the original paper, PC requires knowing the traversed paths of the BFS algorithm, such as the status of each node, either the included ones or \textit{the rejected (and so omitted) ones} in the final returned result. 
Whereas, CoTPC does not require such knowledge, as the CoTs are included as part of the results, not hidden intermediate computations, and the CoT supervision itself can be obtained via the unsupervised discovery method.
Moreover, machine-generated demonstrations can be crowd-sourced and the demonstrators are usually viewed as black boxes, making this a limitation of PC. 

\paragraph{Policy Learning with Motion Planning}
Some existing work \cite{shridhar2023perceiver,james2022q} adopt a strategy similar to CoTPC in terms of predicting key states (the waypoints) as high-level policies.
However, they use motion planners as low-level policies, while CoTPC directly learns to predict low-level controls.
The major advantage of our approach is to handle environments requiring dynamic (or reactive) controls (like Moving Maze and Push Chair), where the key states/observations must be updated at every step and so motion planner-based strategies will struggle. 
Also, PerAct \cite{shridhar2023perceiver} is relatively limited due to its discretized actions, especially for tasks requiring high-precision manipulation (e.g., Peg Insertion).
Moreover, some existing work \cite{qureshi2019motion,fishman2023motion} uses motion planners as the only demonstrators to acquire neural policies, which are to some extent constrained to tasks involving quasi-static control.

\paragraph{Robotic Transformer-1 (RT-1)}
RT-1 \cite{brohan2022rt} is a concurrent work that also directly models low-level control actions with a Transformer.
It benefits from the sheer scale of real-world robot demonstration data pre-collected over 17 months and the tokenization of both visual inputs (RGB images) and low-level actions.
While RT-1 shows great promise in developing decision foundation models for robotics, it adopts the conventional auto-regressive Transformer without explicitly leveraging the structural knowledge presented in low-level control tasks. 
Moreover, it is so computationally intensive that it usually only admits less than 5 control signals per second.
Our work, CoTPC, is an early exploration in this direction and we believe it will inspire the future designs of generally applicable models for robotics tasks.
Another difference is that since RT-1 discretizes the action space, it might suffer from degraded performance for tasks that require high precision (such as Peg Insertion).

\paragraph{Subskill Discovery from Offline Data}
Skill or sub-skill discovery purely from offline demonstration sets is very challenging since there is barely any useful supervision. In the related literature, the option-based approach is a popular strategy. For instance, MO2 \cite{salter2022mo2}, an offline option learning framework using the bottleneck state principle, is shown to work well on continuous control problems for learning the options. However, this line of work has the shortcoming of relying on good state space representation. It is unclear if MO2 can work well for high-dimensional visual observations (e.g., the Pour task only supports visual observation due to soft-body manipulation). Our action-based approach is observation-agnostic and thus avoids this issue. Moreover, the option approach requires learning good initial and termination conditions for the options, which is a hard problem itself \cite{gu2022multi}. Usually, it requires further online learning for skill chaining to compensate for the suboptimal options (MO2 still requires online learning after the options learned from offline datasets are fixed).
The alternative, as we propose, is to utilize methods from the change point detection community for time series modeling to perform action segmentation. Change point detection methods can roughly be divided into predictive model-based approaches (e.g., \cite{saatcci2010gaussian}) and optimization-based ones (e.g., \cite{Killick_2012}). The former requires an underlying predictive model (UPM) where \cite{saatcci2010gaussian} chooses Gaussian Processes. However, it is unclear if it can model high-dimensional and complex control signals as the action sequences can be hard to model in the first place (\cite{saatcci2010gaussian} suggests it might require a large set of sensitive hyper-parameters). The latter does not model the time series directly but finds change points by optimizing a cost function instead. In this case, a good choice of a metric can be critical.

\paragraph{Hierarchical RL + Demonstration regarding Subskill Decomposition}
While the high-level idea of utilizing the hierarchical principles for policy learning is shared with the approaches in the hierarchical RL + demonstration literature (e.g., \cite{jiang2022learning,eysenbach2018diversity,konidaris2012robot,pickett2002policyblocks}), all of these work relies on online interaction with the environments either by using online RL to learn to utilize the learned (sub)skills in a transfer learning setup (e.g, \cite{jiang2022learning}) or by using online exploration to acquire the (sub)skills in the first place (e.g., \cite{eysenbach2018diversity}). On the contrary, ours is a completely offline imitation learning method.
Purely offline subskill discovery that can be used for downstream imitation learning is very challenging, and recent work mostly relies on RL to provide extra supervision \cite{kujanpaa2023hierarchical}. Work that is similar to ours in this setup includes CompILE \cite{kipf2019compile} and PC \cite{yang2022chain}. CompILE adopts a VAE to perform soft-segmentation of the demo trajectories. However, its evaluation was carried out only on a limited set of relatively simple control tasks. On the other hand, PC performs hierarchical policy learning with extra procedure-level supervision that limits its applicability and is hard to be directly compared with. Notice that due to the relatively small previous work in this direction (subskill discovery + hierarchical offline policy learning), both CompILE and PC only use non-hierarchical imitation learning methods as baselines. Similarly in our comparison, we include several non-hierarchical strong baselines.

\section{Details of the Environments}

\paragraph{Moving Maze}
Moving Maze is a 2D maze with a continuous action space of displacement $(\delta x, \delta y)$, where both components $\in [1.5, 4]$.
This an s-shaped maze whose height is 80 and width is 60 with the agent starting from a location randomly initialized inside the top right square region (in green) and the goal is to reach the bottom left one (also in green).
Upon each environment reset, the two regions (the starting square and the target square) as well as the two rectangular bridges (in green) have their positions randomized.
Specifically, The two square regions are randomized to the right of the top and to the left of the bottom blue islands, respectively. 
Their initial locations vary with a range of 20 vertically. 
The two bridges' initial locations vary also with a range of 20 horizontally. 
During the game, each of them except for the top square moves (independently) back and forth with a randomized constant speed $\in [1, 2]$.
Once the agent lands on a moving block, the block becomes static.
The agent cannot cross the borders of the maze (but it will not die from doing so).
For simplicity, we adopt a 10-dim state observation consisting of the current location of the agent and the four green regions. 
This task requires dynamic controls/planning.

\paragraph{Franka Kitchen}
We propose a setting (and thus a variant of the original Franka Kitchen task) where the agent is asked to complete 4 object manipulations (out of 7 different options) in an order specified by the goal.
The 7 tasks are: turn on/off the bottom burner, turn on/off the top burner, turn on/off the light, open/close the sliding cabinet, open/close the hinged cabinet, open/close the microwave oven, and push/move the kettle to the target location.
Compared to the other variants, we use a strict criterion, i.e., the task succeeds when all 4 sub-tasks are completed, where each of them needs to be done in the requested order to be counted as completed.
The environment will terminate when a sub-task other than the specified 4 is performed.
The action space is based on the joint velocity (8-dim) of the robot.
We use the original 30-dim state observation consisting of poses of all the relevant objects and the proprioception signals as well as an additional 14-dim goal embedding.
This embedding assigns a 2-d vector to each of the 7 potential sub-tasks.
Each vector is one of $[0,0],[0,1],[1,0],[1,1]$ (indicating the order to be completed for the corresponding sub-task) and $[-1,-1]$ (meaning the sub-task should not be completed).
We do not include the target pose of the objects in the state observation (i.e., we ask the agent to learn it from the demonstrations).

\paragraph{ManiSkill2}
ManiSkill2 \cite{gu2023maniskill2} is a recently proposed comprehensive benchmark for low-level object manipulation tasks.
We choose 5 tasks as the testbed.
(1) Stack Cube for picking up a cube, placing it on top of another, and the gripper leaving the stack.
(2) Turn Faucet for turning on different faucets.
(3) Peg Insertion for inserting a cuboid-shaped peg sideways into a hole in a box of different geometries and sizes.
(4) Push Chair for pushing different highly articulated chair models into a specified goal location (via a mobile robot).
(5) Pour for pouring liquid from a bottle into the target beaker with a specified liquid level.
All tasks have all object poses fully randomized (displacement around 0.3m and $360\degree$ rotation) upon environment reset (this is in contrast to environments such as Franka Kitchen). 
Note that the holes in Peg Insertion have only 3mm of clearance, requiring highly precise manipulation, and it needs at least half of the peg to be pushed sideway into the holes (in contrast to the similar yet easier tasks \cite{xu2019compare}).
The tasks we select involve both static and mobile manipulation and cover 3 action spaces (delta joint velocity control for Push Chair, delta end-effector pose control for Pour, and delta joint pose control for the rest).
For state observations, we use states of varied dimensions across these tasks (see details in the ManiSkill2 paper) 
For Turn Faucet, we slightly modify the default state observation by appending an extra 3-dim vector (the pose of the faucet link) so that it is easier for the agent to distinguish between different faucet models.
The corresponding demonstrations are modified as well.
For point cloud observations, we use the default pre-processing strategy provided by ManiSkill2 to obtain a fixed length of 1200 points (RGB \& XYZ) per timestep.

\section{Details of the Demonstrations and the Evaluation Protocol}

\paragraph{Moving Maze}
We curate demonstrations by adopting a mixture of heuristics and an RRT-style planner with hindsight knowledge not available at test time.
For each randomized environment configuration, we randomly choose one of the found paths from the starting square to the target one as the demonstration trajectory.
We chunk the maze into 6 regions: the three islands (each bridge belongs to the island below it; the starting square and the target square belong to the top right and the bottom left regions, respectively), each of which is further divided into two regions (cut vertically in the middle).
An RRT-style sampler is used to find paths connecting adjacent regions sequentially (starting from the initial position to the second region).
We restrict the number of steps in each of the paths across two adjacent regions to be  $\le 13$ so that the maximum total length of a demo trajectory is $\le  13\times5=65$ steps.
To enable this type of planning with dynamic environments, we actually first generate demonstrations with a static version of the maze and then animate the moving elements later coherently.
This is not possible during inference time as it requires hindsight knowledge.
We use 400 demo trajectories for training and evaluate all agents on both these 400 configs and a held-out set of 100 unseen environment configs.
We report the task success rate as the major metric.
During inference, we set the maximum number of steps as 60.

\paragraph{Franka Kitchen}
We replay a subset of the human demonstrations originally proposed in \cite{gupta2019relay} in the simulator.
Specifically, we randomly select 50 demo trajectories of length ranging from $150$ to $300$ that succeed in achieving 4 different sub-tasks out of the 7 options.
We relabel them with the privileged information to construct the goal embedding described previously.
Note that this embedding vector is fixed across different time steps for each trajectory.
There are 20 total different ordered sub-task combinations presented in the 50 demonstrations, where the majority of combinations only have $\le 3$ trajectories.
Combinatorial generalization regarding sub-tasks is too challenging in this case (there are $35\times 4!=840$ total combinations); so we focus on evaluating generalization w.r.t. initial robot/object poses.
We use 90 unseen environment configurations (each requiring the completion of 4 sub-tasks) presented in the original human demonstrations for evaluation (we only include seen sub-task combinations).
We report the task success rate (requiring the completion of all 4 sub-tasks in a trajectory) and the average number of successful sub-tasks per trajectory as the metrics.
During inference, we set the maximum number of steps as 280.

\paragraph{ManiSkill2}

For all tasks except for Push Chair, we use the original demonstrations provided by ManiSkill2, which are generated by a mixture of TAMP solvers and heuristics.
Please see the original paper for details (the actual code used to generate these demonstrations is not released, though). 
For Stack Cube and Turn Faucet, we randomly sampled 500 demo trajectories for the training data.
For Turn Faucet, we use trajectories from 10 different faucet models for the demos and perform the evaluation of 0-shot generalization on 4 unseen faucets (each with 100 different scene configurations).
Not that the demonstrations of Turn Faucets have most of the faucets pushed rather than grasped, i.e., under-actuated control.
For Peg Insertion, each different environment config comes with a different shape and size of the box (with the hole) and the peg.
We randomly sampled 1000 trajectories from the original ManiSkill2 demonstrations as the training data and sampled 400 unseen environment configs for the 0-shot generalization evaluation.
For Push Chair, we use a complicated heuristic-based approach from \cite{pan2022silver}, adapted to ManiSkill2 (where it uses additional privileged information and achieves around 50\% task success rate), as the demonstrator.
We collected 1000 successful trajectories as the training data across 6 chair models and we evaluated the 0-shot generalization performance on 300 environment configs distributed over 3 unseen chair models.
For Pour, we use 150 successfully replayed demonstration trajectories provided by ManiSkill2 to generate the point cloud observation sequences (this task does not support state observation as it involves soft-body).
We use the success conditions from the original ManiSkill2 paper to report the task success rate.
During inference, we set the maximum number of steps allowed as 200, 200, 200, 250, and 300 for Stack Cube, Push Chair, Peg Insertion, Turn Faucet, and Pour, respectively.

\section{Illustration of the Discovered Subskill Sequences} \label{app:cos}

\begin{figure}
    \centering   \includegraphics[width=\linewidth]{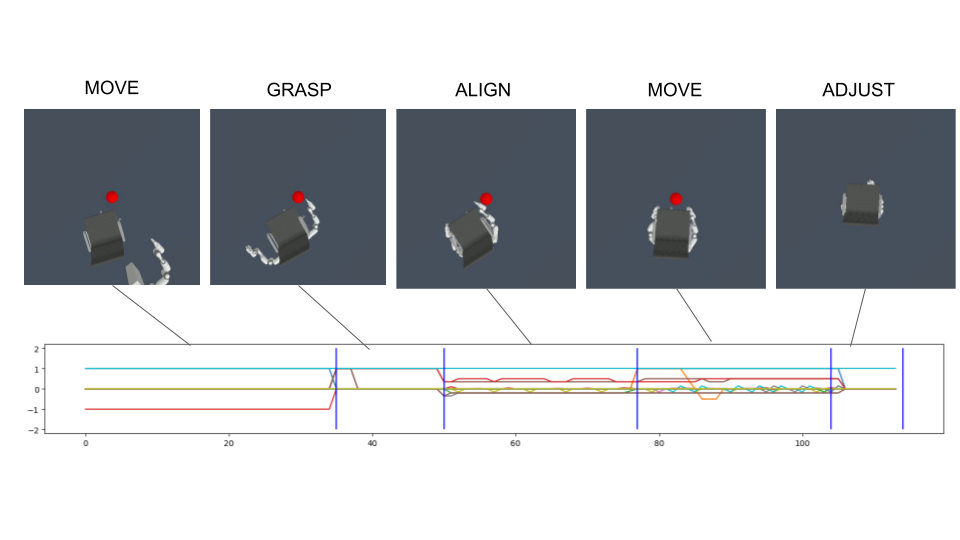}
    \vskip -0.6in
    \includegraphics[width=\linewidth]{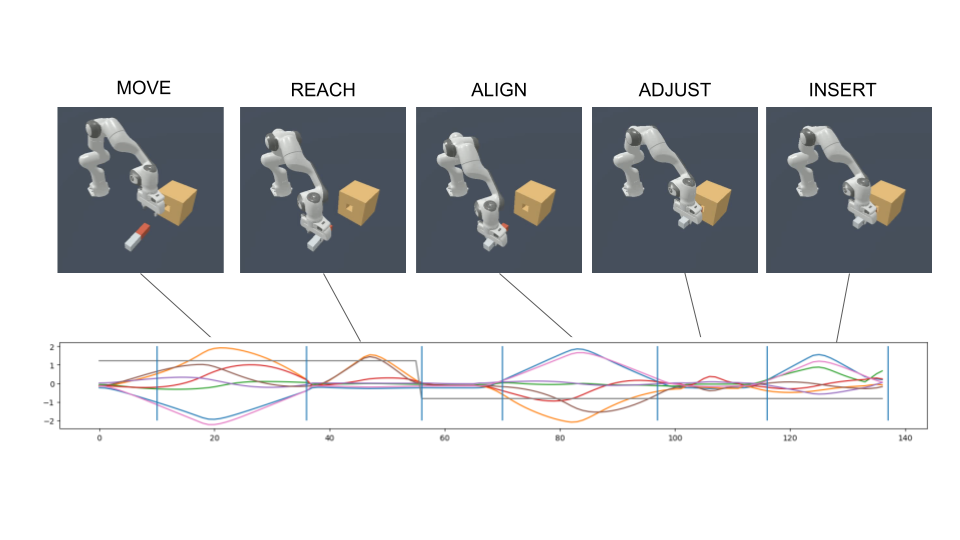}
    \caption{Illustration of actions corresponding to different stages and the associated observations for two tasks: Push Chair (\textbf{top}) and Peg Insertion (\textbf{bottom}). The stages are discovered by grouping the actions into subskills by our unsupervised subskill discovery method.}
    \label{fig:extracted_stages}
\end{figure}

In Fig. \ref{fig:extracted_stages}, we visualize the sub-stages (and thus the subskills) extracted from two trajectories using the automatic subskill discovery process described in Sec. \ref{sec:key_state_selection}.

\section{Implementation Details of Network Architecture and Training}

\paragraph{Vanilla BC}:
We use a Markovian policy implemented as a three-layer MLP with a hidden size of $256$ and ReLU non-linearity.
We train it with a constant learning rate of $1e-3$ with Adam optimizer with a batch size of 32 for 150K iterations (Moving Maze), 300K iterations (Franka Kitchen) and 500K iterations (ManiSkill2).
We find training longer leads to over-fitting even with L2 regularization.

\paragraph{Decision Diffuser}
We use the reference implementation provided by the authors of DD and make the following changes in the diffusion model: 100 diffusion steps, 20 context size, and 4 horizon length (in our experiments we found that longer performs worse). The diffusion and inverse-dynamics models have $\sim$1.6M parameters in total.
Since DD works on fixed sequence lengths, we pad the start and end states during training and only the start states during inference.

\paragraph{Diffusion Policy}
We use the reference implementation provided by the authors of DP while adjusting the model size to be $\sim$1.2M parameters (comparable to other baselines and our CoTPC).


\paragraph{Decision Transformer}
We adopt the \href{https://github.com/karpathy/minGPT}{minGPT} implementation and use the same set of hyperparameters for all tasks (a feature embedding size of 128 and 4 masked multi-head attention layers, each with 8 attention heads), totaling slightly greater than 1M learnable parameters.
The action decoder is a 3-layer MLP of two hidden layers of size $256$ with ReLU non-linearity (except that for Franka Kitchen we use a 2-layer MLP of hidden size $1024$).
We train DT with a learning rate of $5e-4$ with a short warm-up period and cosine decay schedule to $5e-5$ for all tasks (except for Franka Kitchen, whose terminal lr is $5e-6$ ) with the Adam optimizer with a batch size of $256$.
We train for 200K iterations for Moving Maze and Franka Kitchen and train for 500K iterations for all tasks in ManiSkill2.
We use a weight decay of $0.0001$ for all tasks but Franka Kitchen (for which we use $0.1$) and Push Chair (for which we use $0.001$). 
We use a context size of 10 for ManiSkill2 tasks, 20 for Moving Maze, and 10 for Franka Kitchen.
We use learnable positional embedding for the state and action tokens following the DT paper.

\paragraph{Behavior Transformer} 
In this paper, we use the modified version of BeT where a context of both past actions and past states is used to predict the actions.
We started with the configuration used for the Franka Kitchen task in the original paper.
We changed the number of bins in K-Means to 1024 (we find that for our tasks, a smaller number of bins works worse) and changed the context size to 10 (in line with the other transformer-based models).
The Transformer backbone has approximately the same number of parameters ($\sim$1M) as CoTPC and DT.
We train the model for around 50k iterations (we find that training longer leads to over-fitting easily for BeT, potentially because of its discretization strategy and the limited demos used for BC).
For ManiSkill2 tasks, we use the same architecture as that for DT, except that we use the center plus offset decoders to decode the actions.
We use a k-means of $128$ clusters to partition the action space.
We train 100K iterations for Turn Faucet, Push Chair, and Pour. 
We train 200K iterations for Peg Insertion and Stack Cube.
We use a coefficient of 100 for training the action offset regressor (i.e., the action offset regressor loss is scaled up 100 times compared to the action center classification loss).
Unless otherwise specified, we follow all the other hyperparameters described for DT.

\paragraph{CoTPC}
Unless specified here, we keep other configurations (both model training and network architecture) the same as those in BeT.
We use no positional embeddings for CoT tokens as they themselves are learnable prompts. 
The CoT decoder is a 2-layer MLP with ReLU non-linearity of hidden size $256$.
We use a coefficient $\lambda=0.1$ for the auxiliary MSE loss for all tasks.
During training, we apply random masking to the action and state tokens so that the CoT tokens attend to a history of varied length (from the first step to a randomized $t$-th step). Also, see the main paper for more details.
For all tasks, we use the same number of clusters to partition the action space as for the BeT baseline.
Another implementation detail: we find that using CoT features from other than the last attention layer $\textbf{S}_{j}^{\textbf{cot}}(\cdot), j \le J$ as inputs for the CoT decoder can further improve the final performance. 
We consider this as a per-task hyper-parameter.
In general, we find that using $\textbf{S}_J^{\textbf{cot}}(\cdot)$ is a good starting point.
Unless otherwise specified, we follow all the other hyperparameters described for BeT.

\section{Choice of Metric for Unsupervised Subskill Discovery}
\label{app:metric}
Our subskill discovery approach by decomposing the action sequences is motivated by the intuition that functionally similar actions that are temporally close should be grouped into the same subskill. The choice of the metric to measure such functional similarity is an open problem. We tried cosine similarity, L2 distance, and a more complicated Hausdorff distance. Among these, we find that the cosine similarity exhibits both simplicity and generalizability. In our experiments, we find it works across different action spaces (e.g., delta position in Moving Maze, delta joint pose in Peg Insertion, and delta joint velocity in Push Chair) with different action statistics based on how they are generated (sampling-based methods such as RRT, heuristics, etc.) and scale well to higher action dimensions (e.g., Moving Maze uses 2-d actions, Push Chair uses 19-d actions with a dual-arm mobile robot). In practice, action spaces of much higher dimensions are relatively uncommon in robotic tasks.

\section{Hyper-parameter Tuning for Granularity of the Subskill Discovery}
\label{app:beta}

We tune $\beta$ in the subskill discovery process, which controls how granular we want for the discovered subskill sequences.
For different action spaces (e.g., dual-arm vs. single-arm), we visualize a small set of 5-10 demo trajectories as the validation set to tune $\beta$. 
We find $\beta$ relatively insensitive with $0.05$ for all single-arm control tasks and $0.02$ for all dual-arm control tasks. 
Moreover, our proposed model is relatively robust regarding $\beta$. 
In an ablation study, we manually tuned it so that the average length of the discovered subskills varies. 
We found that on the Peg Insertion task (with state observation policy setup), approximately one subskill shorter (on average) leads to slightly decreased performance (59.3 vs. 54.0 in terms of 0-shot SR), and similarly for approximately one subskill longer (59.3 vs. 55.3 in terms of 0-shot SR). 
With a significantly longer length of the discovered subskills (around 7), however, the performance drops significantly, since in this case, the overly segmented subskills do not provide much valuable structural information in policy learning.

\begin{figure}
    \includegraphics[width=\linewidth]{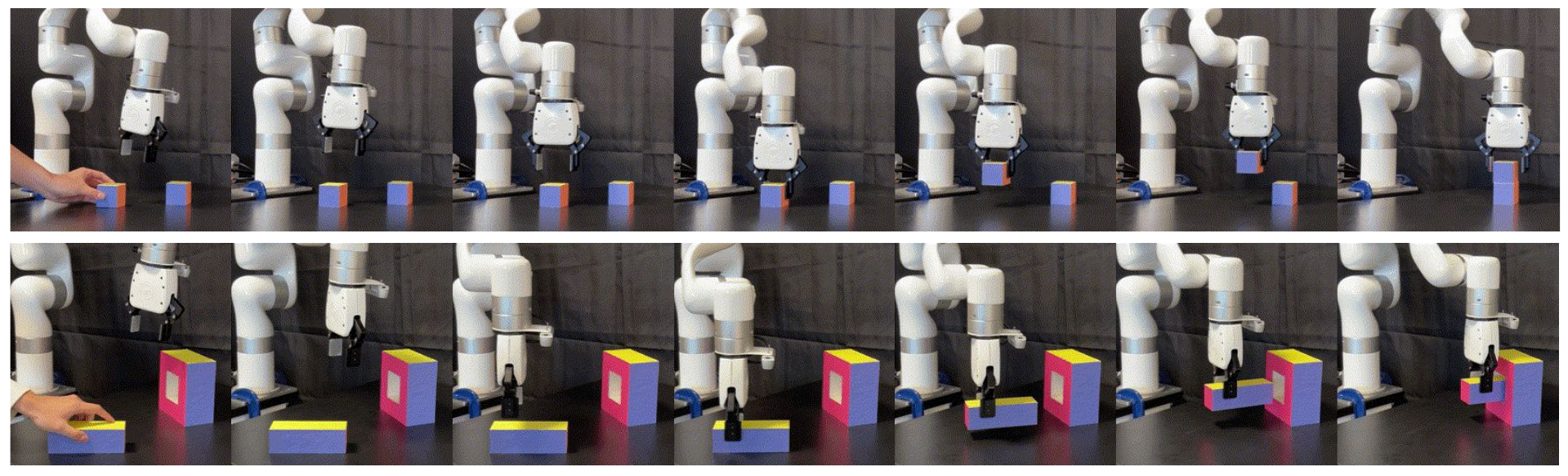}
    \caption{Two sampled succeeded trajectories for Stack Cube and Peg Insertion, respectively, in a real robot setup, from state-based CoTPC policies trained purely from demos in simulators.
    As an early examination, we increase the clearance for peg insertion from 3mm (sim) to 10mm (real) and only use peg and box-with-hole models of fixed geometry.}
    \label{fig:real}
\end{figure}

\section{Preliminary Results of Sim-to-Real Transfer} \label{app:s2r}

We examine the plausibilities of sim-to-real transfer of our state-based CoTPC in a zero-shot setup on two tasks, namely, Stack Cube and Peg Insertion.
Our real-world experiment setup and two sampled succeeded trajectories are illustrated in Fig. \ref{fig:real}.
With an off-the-shelf pose estimation framework such as PVNet \cite{peng2019pvnet}, we can achieve reasonable performance using the state-based CoTPC policy learned purely from simulated data.
See videos on our project \href{https://sites.google.com/view/cotpc}{page}. 
As a preliminary examination, we only perform qualitative evaluations.

\section{Additional Results on ManiSkill2 (Mean \& Standard Deviation Among 3 Runs)} \label{app:var}
In addition to the main paper's results where we report the best among the three training runs in the original paper (similar to the eval protocol in \cite{chi2023diffusion}), we also report the mean and std in Tab. \ref{tab:more_main_results_}. The format is best (mean $\pm$ std).

\begin{table*}[t]
\caption{Additional test performance (success rate) on the unseen and the 0-shot setup for ManiSkill2 tasks with state observations.}
\vskip 0.1in
\label{tab:more_main_results_}
\begin{center}
\begin{small}
\begin{sc}
\begin{tabular}{l c c c c}
\toprule
  & Stack Cube & Peg Insertion & Turn Faucet & Push Chair \\
\midrule
 & {\scriptsize unseen} & {\scriptsize 0-shot} & {\scriptsize 0-shot} & {\scriptsize 0-shot} \\
\midrule
Behavior Transformer & 73.0 (68.7$\pm$4.0) & 42.5 (39.5±2.7) & 32.5 (31.7±0.8) & 33.4  (32.3±1.0) \\
CoTPC (ours) & 86.0 (84.0±2.6) & 59.3 (51.5±7.4) & 39.3 (35.4±3.6) & 41.0 (36.4±4.0)\\
\bottomrule
\end{tabular}
\end{sc}
\end{small}
\end{center}
\end{table*}
\vskip -0.05in

\section{Details of Point Cloud-based CoTPC} \label{sec:pc_CoTPC}

\begin{figure}
    \includegraphics[width=\linewidth]{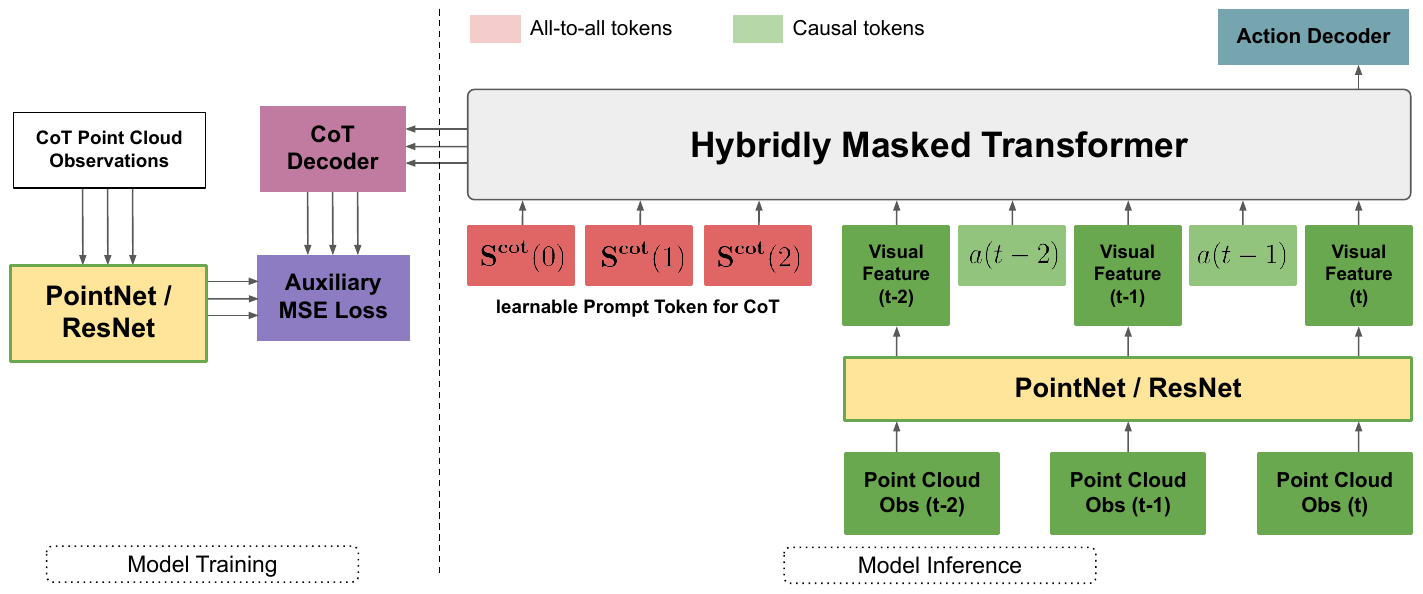}
    \caption{Illustration of the point cloud-based CoTPC. Compared to state-based CoTPC, we add a PointNet to process the point cloud observations as well as the point cloud CoTs. We omit the data path for the input proprioception signals to the model. We also omit the detailed data flow for the action decoders.}
    \label{fig:pc_CoTPC}
\end{figure}

To process point cloud observations, we adopt a lightweight PointNet \cite{qi2017pointnet} ($\sim$27k parameters) that is trained from scratch along with the transformer in an end-to-end manner. 
We concatenate additional proprioception signals with the point cloud features to the input state tokens.
In CoTPC where the CoT decoder is trained to predict the point cloud CoT, we ask the decoder to predict PointNet features of the CoT instead.
We find that the auxiliary point cloud CoT loss causes the PointNet encoded representations to collapse. Inspired by \cite{chen2021exploring}, we use a stop-gradient operation in the point cloud CoT encoding path to prevent this.
We illustrate the network architecture in Fig. \ref{fig:pc_CoTPC}.
The training strategies (we use the same set of hyperparameters) and evaluation protocols are similar to those of the state-based experiments.

\section{Additonal Videos}
We provide videos (.gif animations) of inference results for ManiSkill2 tasks from our state-based CoTPC on the project \href{https://sites.google.com/view/cotpc}{page}.


\end{document}